\documentclass[letterpaper, 10 pt, journal, twoside]{IEEEtran}
\usepackage[pdftex]{graphicx} 
\usepackage{gensymb}
\usepackage{amsmath}
\usepackage{hyperref}
\IEEEoverridecommandlockouts  
\usepackage{booktabs}
\usepackage{xcolor}
\usepackage[normalem]{ulem}
\usepackage{tikz}
\usepackage{xfrac}
\usepackage{float}
\usepackage{placeins}
\usepackage[justification=centering]{caption}
 \usepackage[pscoord]{eso-pic}
\begin{document}

\title{Deep Learning for Polycystic Kidney Disease (PKD): Utilizing Neural Networks for Accurate and Early Detection through Gene Expression Analysis}
\author{
\begin{tabular}{c} Kapil Panda \\ University of North Texas \\ Denton, United States of America \\ kapil.panda30sc@gmail.com \end{tabular} \and
\hspace{50pt} 
\begin{tabular}{c} Anirudh Mazumder \\ University of North Texas \\ Denton, United States of America \\ anirudhmazumder26@gmail.com \end{tabular}
}
\maketitle
\begin{abstract}
With Polycystic Kidney Disease (PKD) potentially leading to fatal complications in patients due to the formation of cysts in kidneys, early detection of PKD is crucial for effective management of the condition. However, the various patient-specific factors that play a role in the diagnosis make it an intricate puzzle for clinicians to solve, leading to possible kidney failure. Therefore, in this study we aim to utilize a deep learning-based approach for early disease detection through gene expression analysis. The devised neural network is able to achieve accurate and robust prediction results for possible PKD in kidneys, thereby improving patient outcomes. Furthermore, by conducting a gene ontology analysis, we were able to predict the top gene processes and functions that PKD may affect.
\end{abstract}


\section{Introduction}
Polycystic Kidney Disease (PKD) is a prevalent, yet under-researched hereditary renal disorder characterized by the formation of numerous cysts in the kidneys \cite{Bergmann2018}. Over time, these cysts can enlarge and disrupt the standard kidney structure, impairing kidney function and thus leading to various complications, such as high blood pressure, kidney stones, and, in severe cases, kidney failure \cite{CKD}. Even though this disease is regarded as one the top kidney ailments of patients in the US, it is still relatively unexplored, hence leading to late detection and ineffective care. Therefore, early and accurate diagnosis of PKD is crucial for timely intervention and effective management of the condition to improve patient outcomes, however, the various patient-specific factors that play a role in the diagnosis make it an intricate puzzle for nephrologists and clinicians to solve \cite{cyst}. 

In recent years, the progress made in artificial intelligence and machine learning, specifically deep learning, has opened up new possibilities in healthcare for both detection and prediction \cite{DL}. Neural networks, a cornerstone of deep learning, have demonstrated exceptional proficiency in image recognition, feature extraction, and classification tasks \cite{NN}. With the incredible capabilities of these algorithms and models, healthcare professionals now have powerful tools that can aid in analyzing vast amounts of medical data with unparalleled precision and efficiency, thereby increasing the efficacy of medical prescriptions. However, certain domains have yet to be immersed in the frontiers of AI and ML, with PKD being one such disease that has had no previous research using machine learning done in it\cite{Panda2023}.

Therefore, in this research we aim to leverage the power of deep learning by utilizing neural networks to aid in PKD detection for accurate and early diagnosis. Utilizing methods such as synthetic data creation and data preprocessing, an MLP algorithm and stacking ensemble were trained to see if they could learn whether or not a patient had PKD. Furthermore, using a gene ontology tool, we found robust results indicating the processes and functions of the gene expressions that the model found highly correlated with PKD to gain deeper insights into the underlying molecular mechanisms that are affected by the disease.

\section{Methodology}
\subsection{Materials}
The materials that were used for this project were Python for the computation and mice gene expression data that was acquired from \cite{Chen2008}.

\subsection{Mice Data Based Algorithm}
\subsubsection{Data Preprocessing}
The algorithm started with the mice-based data getting pre-processed, allowing an algorithm to be used to learn the data. Due to limitations on acquiring human data on PKD, we had to resort to mice genetic data. According to \cite{Ernst2018}, almost all of the genes in mice share some functions with the genes in humans, especially in kidneys, allowing for mice to be indicative test subjects on the impact of PKD on humans. Thus, it is not uncommon in the industry for laboratories to use mice data when testing for human diseases. 

Within this dataset, some of the data was dropped to eliminate all the undefined points in the dataset and then a scaler was used to standardize the data. 

\subsubsection{Machine Learning}
A Multilayer Perceptron Classifier(MLP) was utilized to learn the data. An MLP was utilized due to their ability to solve nonlinear problems. They are good at learning the relevant features and data, and they stack a layer of their neurons, where each layer is learning different parts of the data, allowing the neural network to get a hierarchical representation of the data. Additionally, MLPs are based on backpropagation, in which they can update the weights of different connections between neurons to minimize the difference between the actual and predicted output values. 

An MLP was used for this data rather than a stacking ensemble due to its ability to understand the complexity of data better than stacking ensembles. In contrast, stacking ensembles are better at highlighting the diversity of the models, working better than pure individual models. Additionally, the outputs of MLPs are more complex to understand because they are neural networks rather than machine learning algorithms. Lastly, MLPs are better able to capture an understanding of the data than stacking ensembles because MLPs are based on neural networks. In contrast, stacking ensembles are based on smaller, more basic models, all utilized together.

\subsubsection{Clustering Algorithm}
The data was clustered based on the probability of getting PKD and the gene expression. To cluster, a K-means algorithm was used because of its ability to maximize the similarity of variables within each cluster while also minimizing the similarity of variables in different clusters, which ensures that when it is creating the centroids, they are as distinct as possible from one another, while also being very indicative of what data belongs in each cluster. Additionally, based on the output of the clustering algorithm, it can be further interpreted to see what genes are most indicative of PKD. When the clustering algorithm was created, three centroids were utilized to cluster the data to create specific groupings of which genes had the highest likelihood of getting PKD compared to the other genes.

\subsection{Synthetic Data Based Algorithm}
\subsubsection{Data Creation}
In this study, we also used syntehtic data for our model. The utilization of synthetic data in this research is of paramount importance for several reasons \cite{James2021}. Firstly, it addresses the inherent challenge of limited real-world clinical data on PKD. Synthetic data serves as a crucial supplement, enabling the development and training of machine learning algorithms in the absence of comprehensive clinical datasets. Secondly, synthetic data creation provides an opportunity to control and manipulate various parameters, such as the percentage of patients with PKD, which is challenging with real clinical data. This flexibility allows for comprehensive experimentation and model training under different scenarios, ultimately enhancing the robustness and adaptability of the algorithm.

The synthetic data was created through Sklearn's synthetic dataset creation algorithm. Synthetic data is generated by a model rather than using real-world data. This data creation method was utilized for our problem due to a need for real-world data about humans with PKD \cite{jordon2022synthetic}. The parameters that were used to create this data were 1000 samples were created, 5000 features were created with 100 of the features being important, two classes were created(has PKD or not), and the values were weighted at 20\% chance of having PKD. All of these values were kept as controls throughout the research. However, for future analysis, the 20\% value could change based on further research on the percentage of patients who have PKD. Synthetic data is more accessible for the data augmentation needed to see if a machine learning algorithm could learn whether or not a patient has PKD based on gene expression data due to the data for PKD being hard to collect and analyze. Additionally, using this specific set of parameters, a full synthetic dataset was created to allow the machine learning model to be trained to evaluate further if it could accurately understand how to classify whether or not a patient has PKD.
\subsubsection{Data Preprocessing}
Data preprocessing was done to use the dataset and to split the data. A scaler was used to standardize the data by removing the mean and scaling to the variance. Additionally, scaling the data was vital to ensure that the dataset's features were standardized, which is useful when dealing with complex features with varying scales and distributions. Additionally, this helps machine learning algorithms converge more and converge faster due to their ability to minimize the difference in feature scales. Then, a train test split happened, splitting the data into 80\% training data and 20\% testing data.

\subsubsection{Machine Learning}
The machine learning algorithm had a few parts, as it was a stacking ensemble-based machine learning algorithm. The stacking ensemble consisted of three different machine learning algorithms to learn the data: Support Vector Classifier(SVC), Random Forest Classifier(RF), and Gradient Boost Classifier(GB). The meta-classifier was a Logistic Regression(LR) algorithm. An SVC was used due to its ability to separate between classes and because it can find complex relationships within the data. Additionally, an RF was used as part of the machine learning algorithm due to its ability to achieve high accuracy and robustness and distinguish which features were the most important. Lastly, a GB was used due to its ability to provide high accuracy while focusing on decreasing the error due to bias within the algorithm. Additionally, an LR was utilized for the machine learning algorithm as the meta-classifier as it allows the creation of weights to the prediction on the base models, allowing for an aggregation that considers individual processors' reliability.

On top of all the independent machine learning algorithms, a stacking ensemble was used to accurately utilize all of the algorithms to create a more robust machine learning model. A more robust model is created by utilizing all of the strengths of the independent machine learning algorithms to accurately predict if a patient has PKD based on synthetic data. These strengths could also be weighted based on the LR algorithm, as it can show which base models have the most significant effect on the machine learning algorithm and how it can maximize the accuracy. Additionally, the stacking ensemble allows for all of the algorithms to be able to make their predictions. In contrast, the stacking ensemble can utilize the other algorithms' responses and curate the best response. The ability of the stacking ensemble to curate the best response allows it to achieve high accuracies and figure out what parts and pieces of data are the most indicative when predicting.

\section{Results}
\subsection{Model Performance on Synthetic Data}
The developed machine learning model achieved an accuracy of 78\% when run on the synthetic data produced, indicating a moderate success level. However, this level of accuracy needs to be improved for a diagnostic tool intended for PKD detection. In the context of PKD, where early and accurate diagnosis is crucial for effective intervention and management, a sub-optimal accuracy level like this could lead to misdiagnoses or delayed treatments, potentially exacerbating patient outcomes.
\FloatBarrier
\begin{figure}[!htb]
	\centering
	\includegraphics[width=1\columnwidth]{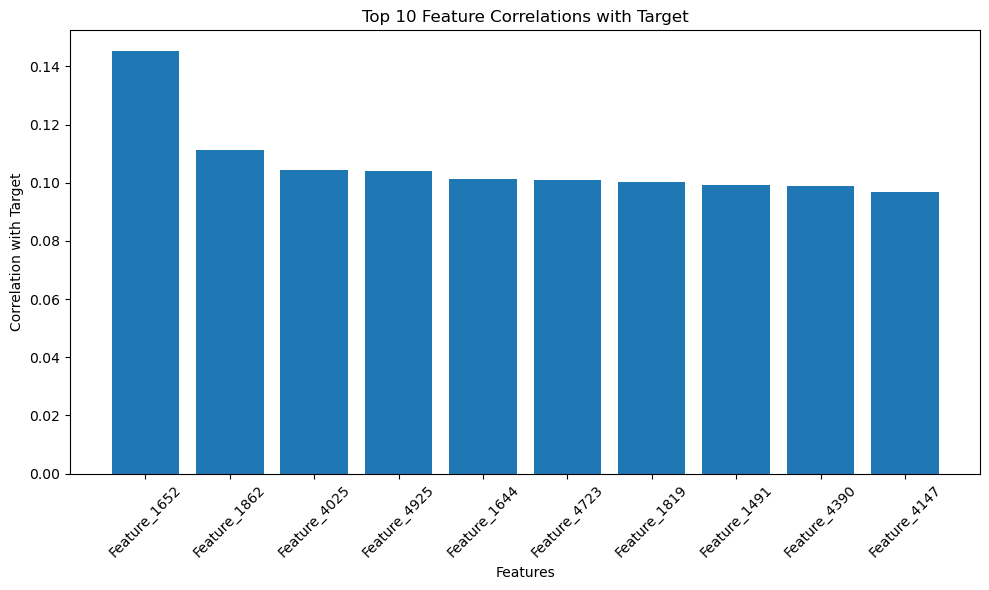}
	\caption{Feature Correlations on Synthetic Data}
	\label{fig: Figure 1}
\end{figure}
\FloatBarrier

\subsection{Model Performance on Mice Data}
A more realistic dataset was used to address the initial model's performance limitations on synthetic data acquired from mice afflicted with PKD. Remarkably, upon testing the model using this more representative dataset, an accuracy of 92.23\% was achieved, signifying a remarkable leap in predictive performance. The substantial improvement in accuracy suggests that the Multilayer Perceptron Classifier (MLP) successfully leveraged the realistic data to capture the disease's intricate patterns. This achievement is particularly significant as it points toward the model's potential to discern more distinct and nuanced data representations, which is crucial when dealing with complex conditions such as PKD.
\FloatBarrier
\begin{figure}[!htb]
	\centering
	\includegraphics[width=1\columnwidth]{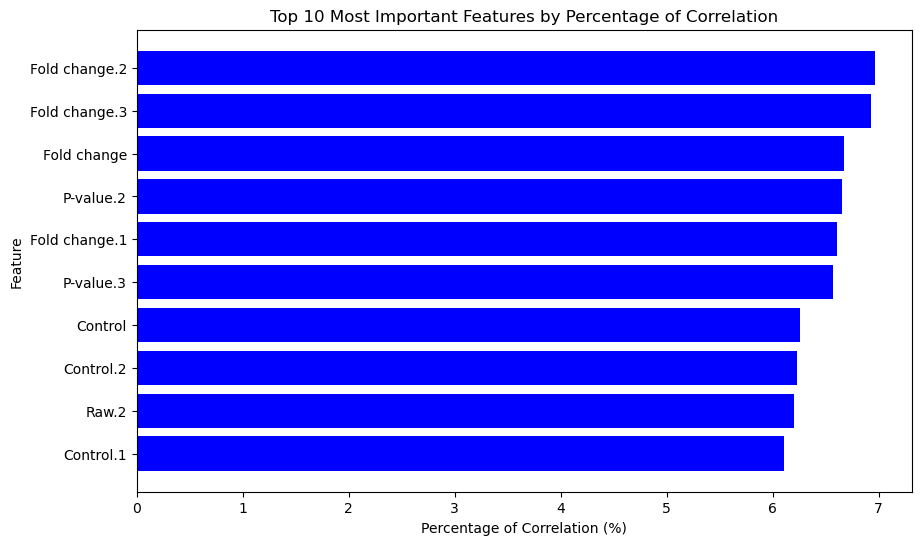}
	\caption{Feature Correlations on Mice Data}
	\label{fig: Figure 2}
\end{figure}
\FloatBarrier
\subsubsection{Feature Selector}
The feature selector shows which features had the highest correlation with whether or not the mouse had PKD. Although the values are relatively low, they can definitively identify the possible features leading to the MLP algorithm accurately predicting whether or not a mouse has PKD. As seen in \ref{fig: Figure 2}, the algorithm is affected by the fold changes in the gene expression the most, and it shows the highest correlation with the computed output, showing potential for the fold change to identify a relationship with patients with PKD. Although the fold change is the most indicative, it is relatively close to the over features, showing that many of these different features impact the MLP's understanding of the data and its ability to classify whether or not a mouse has PKD.

\subsection{Comparison}
\subsubsection{Dataset}
Some key differences potentially lead to the difference in accuracy between the two different algorithms. One significant difference between the datasets of the two different algorithms is that synthetic data is generated data, so it may be that the data being created is not indicative of anything particularly related to PKD, making it more prone to lowered accuracy than the standard clinically tested mice data. On the other hand, although the synthetic data may not be helping the algorithm, it is possible that the mice data needs to be more generalizable to humans due to the differences between the two species. Additionally, this shows that it is vital to see whether or not it is possible to make synthetic data contextual to humans, utilizing more popular resources such as mice data.

Although, there is a difference between mice and human data, the mice data still has indications on the human processes. As seen in \cite{Perlman2016}, mice are very good for modelling biological processes in humans, as they have a lot of biological processes which are similar. Additionally, specifically in the context of kidneys, there is evidence that mice have similar kidney processes as humans.

Additionally, due to a lack of adequate human clinical data on polycystic kidney disease (PKD), the research team had to rely heavily on introspection studies of PKD in mice models as well as synthetic data generated from computational modeling. Without access to sufficient real-world data from human PKD patients to train machine learning algorithms, the use of these alternative data sources like mouse models and synthetic data was absolutely vital. The mouse models, while not a perfect analogue, still provided important insights into the biological mechanisms and progression of PKD that could inform the development of AI systems. The synthetic data served as a supplementary training set to "fill in the gaps" where human clinical data was missing or inadequate for properly training the algorithms. Without the ability to leverage these alternate forms of data, it would have been tremendously difficult to develop AI systems capable of reliably analyzing and understanding the complexities of human PKD cases. The multi-modal approach, combining real human data where available with data from mouse models and synthetic data generation, provided a robust overall training dataset that allowed the research team to make meaningful progress in applying AI to better understand PKD.
\subsubsection{Algorithms}
The different models may be leading to a difference in the accuracy of the two algorithms. The difference could be because MLP is a neural network, potentially allowing it to learn and understand the patterns in the data better. At the same time, stacking ensembles are bogged down to the multiple different, more general, and simple base classes. On the other hand, the stacking ensemble may be more robust since it is built with a few other methods on top of the actual stacking-based model.

\subsection{Clustered Analysis}
\begin{figure}[!htb]
	\centering
	\includegraphics[width=\columnwidth]{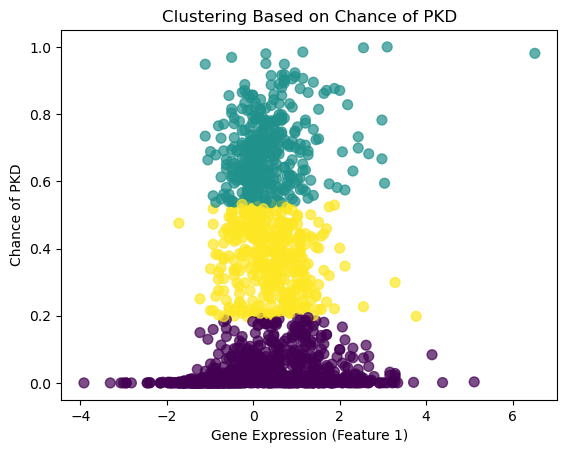}
	\caption{Clustering Based on Chance of PKD}
	\label{fig: Figure 3}
\end{figure}

The results of the clustering algorithm can be seen in Figure \ref{fig: Figure 3}. The results of the clustering algorithm show which gene expression is the most likely to have PKD. By looking at this data set, we can centralize the genes to see which ones are most likely to get PKD, further allowing us to analyze this gene expression using a gene ontology tool. 
\FloatBarrier
\FloatBarrier
\subsection{Gene Ontology}
\begin{figure*}[!htb]
	\centering
	\includegraphics[width=1.35\columnwidth]{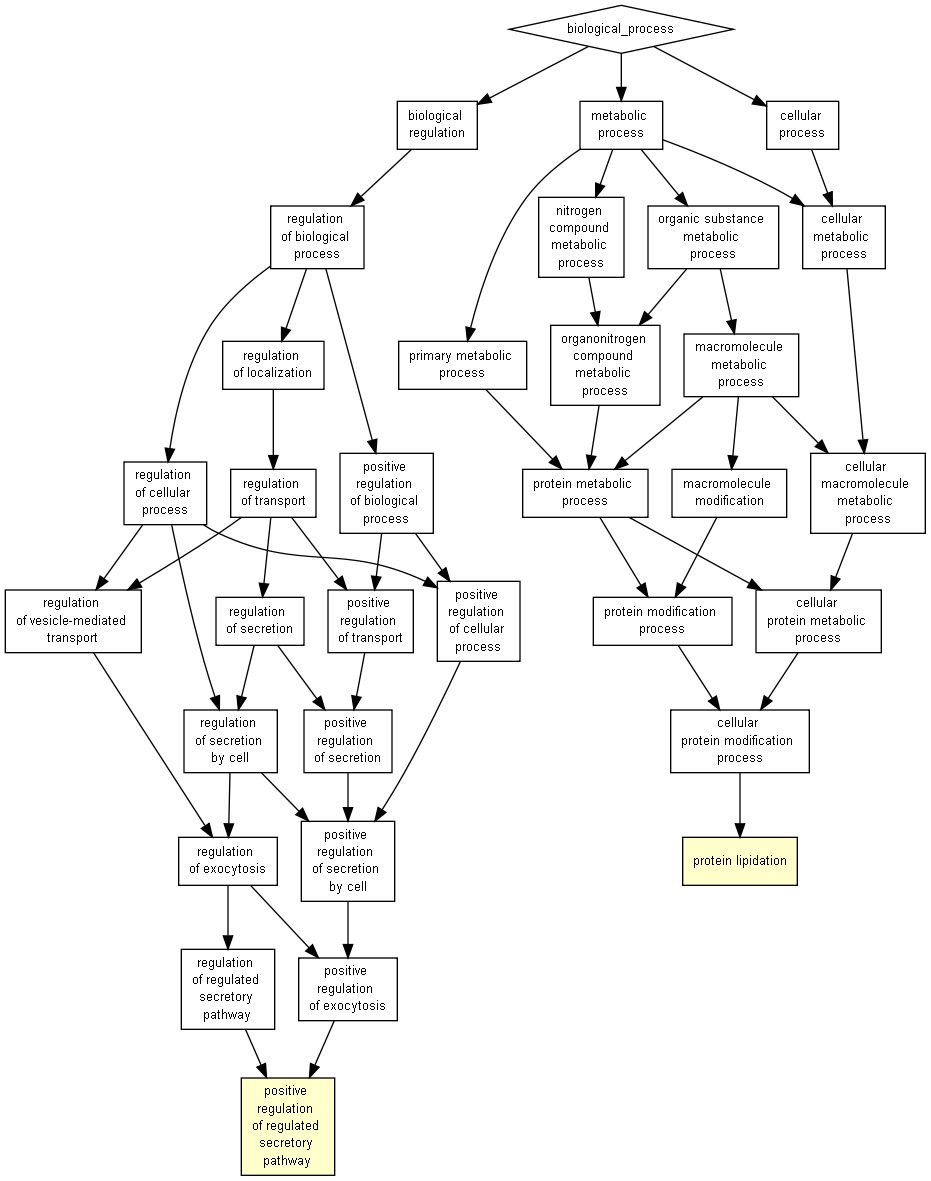}
	\caption{GOProcess Gene Process Analysis}
	\label{fig: Figure 4}
\end{figure*}

To better understand the genes most indicative of PKD \cite{aki}, we focused on the top cluster identified by our model since it showed the highest risk of getting PKD and conducted a gene ontology analysis using the GoProcess enrichment tool \cite{Eden2009GOrilla}.
A comprehensive analysis of the clustered gene expressions yielded intriguing insights into the molecular underpinnings of PKD, shedding light on potential mechanisms that contribute to the disease's pathogenesis. As depicted in Figure \ref{fig: Figure 4}, our investigation revealed a distinct enrichment of gene processes that are particularly affected within the top cluster. Notably, two prominent gene processes, "Positive Regulation of Regulated Secretory Pathway" \cite{Lin2013} and "Protein Lipidation" \cite{Jiang2018}, emerged as focal points for understanding the complex interactions driving PKD.

Positive regulation of regulated secretory pathways encompasses a cellular phenomenon characterized by the augmentation of a precise and controlled process involving releasing specific molecules from cells. This process is orchestrated in response to intricate signaling cues, highlighting the intricacies of cellular communication. Our findings indicate a noteworthy connection between PKD-associated genes in this cluster and their potential influence on augmenting the release of specific molecules from cells. This heightened secretion, triggered by the upregulated gene processes, could play a pivotal role in the disease's progression, potentially implicating aberrant cellular communication pathways in developing cystic structures within the kidneys.

\begin{figure}[!htb]
	\centering
	\includegraphics[width=0.4\columnwidth]{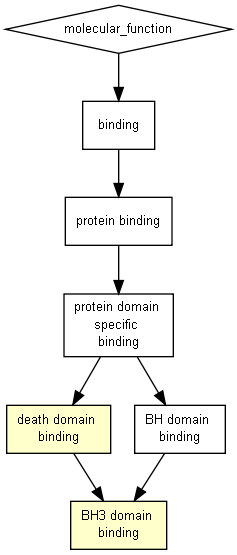}
	\caption{GOProcess Gene Function Analysis}
	\label{fig: Figure 5}
\end{figure}

Another correlating process our study found to PKD-related genes within the top cluster was that of protein lipidation. This process involves attaching lipid molecules, often fatty acids, to proteins. This modification can significantly impact proteins' structure, function, and localization within cells. Our findings suggest that the genes linked to PKD in this cluster may influence the lipidation of proteins, potentially leading to altered cellular functions. The intricate interplay between lipids and proteins is a fundamental aspect of cellular physiology, and its disruption could contribute to the anomalies observed in the context of PKD.
\FloatBarrier
\FloatBarrier
Furthermore, in this analysis, we aimed to identify the gene functions most affected by PKD in this top cluster. As seen in Figure \ref{fig: Figure 5}, we found a distinct enrichment of the gene functions "BH3 Domain Binding" \cite{Chittenden2002} and "Death Domain Binding" \cite{SINGH1998} when analyzing the gene ontology.

BH3 domain binding is a fundamental aspect of cellular regulation, particularly in cell survival and apoptosis. The BH3 domain, an essential structural motif in specific proteins, is central in orchestrating cellular responses to stress signals and external cues. Our findings suggest a profound connection between PKD-associated genes within the identified cluster and their involvement in BH3 domain binding interactions. This observation offers a tantalizing glimpse into the potential regulatory mechanisms that these genes might modulate, possibly influencing the delicate balance between cell survival and programmed cell death. The aberrations in BH3 domain binding interactions might contribute to the abnormal cellular processes associated with PKD, hinting at a potential avenue through which the disease exerts its effects.

Similarly, our exploration of death domain binding interactions within the context of PKD uncovers an intricate layer of molecular signaling pathways that might be perturbed in disease progression. The death domain, a distinct protein module implicated in transmitting signals related to cell death and inflammation, is crucial in orchestrating cellular responses to various stimuli. Identifying PKD-associated genes within this cluster as influential in death domain binding interactions signifies their potential involvement in shaping the cellular fate decisions that underpin PKD's development. The disruption of these interactions could contribute to the misregulation of cell survival and inflammatory responses, both of which are pivotal aspects of PKD pathology.
\FloatBarrier
\FloatBarrier 

\section{Discussion}
\subsection{Conclusion}
This research devised an effective and accurate prediction model for early detection of Polycystic Kidney Disease with the help of deep learning. While the initial accuracy of 78\% achieved on synthetic data marked a promising starting point, the significant leap to an accuracy of 92.23\% when employing the MLP algorithm on real mice data signifies a remarkable advancement in PKD detection. Furthermore, exploring clustering techniques for gene expression data further enriched our understanding of PKD, allowing us to identify gene clusters with higher likelihoods of PKD occurrence. The subsequent gene ontology analysis offered glimpses into specific processes and functions influenced by PKD-associated genes, shedding light on potential pathways and molecular interactions involved in PKD development and progression, as well as possible effects of PKD on the body.
\subsection{Potential Implications}
The findings of this study carry significant implications for the field of Polycystic Kidney Disease (PKD) detection and diagnosis, as well as the broader intersection of artificial intelligence and healthcare. The exploration of machine learning techniques in PKD diagnosis underscores the potential of advanced computational methods to address complex medical challenges.

The achieved accuracy of 78\% on the synthetic dataset serves as a valuable insight into the initial capabilities of machine learning models in PKD detection. While this accuracy signifies a notable advancement, its limitations are apparent in the context of PKD's critical need for accurate and early diagnosis. However, the remarkable leap in accuracy to 92.23\% achieved with the MLP algorithm using the mice dataset presents a promising avenue for improving PKD detection. The success of the MLP algorithm in capturing the nuances of PKD patterns reflects its potential to discern complex data representations, critical for a condition as multifaceted as PKD. This achievement highlights the power of deep learning models in delving deeper into data intricacies, and it encourages further exploration into utilizing neural networks for medical diagnostic tasks. Furthermore, 

On the other hand, comparing the performance of the MLP algorithm and the stacking ensemble offers insights into the strengths and limitations of different machine learning approaches. The substantial difference in accuracy raises questions about the fundamental differences between these algorithms in terms of their ability to understand complex relationships within data. This comparison also prompts a deeper examination of the underlying mechanisms that contribute to the MLP's outstanding performance and whether similar capabilities could be integrated into other ensemble-based models.

Finally, the study's methodological insights offer considerations for future research directions. The potential limitations of synthetic data and the need to validate models on authentic human data underscore the importance of diverse and representative datasets. The exploration of clustering methods for gene expression data has potential implications beyond PKD diagnosis, providing a means to identify gene clusters that could inform further research into disease mechanisms and potential therapeutic targets. In the context of healthcare, this research showcases the evolving role of artificial intelligence in medical diagnosis. While this study focuses on PKD, the methods and insights can serve as a template for applying machine learning techniques to other complex diseases, where early detection and precise diagnosis are of paramount importance. This research thus contributes to the ongoing dialogue on the integration of AI and healthcare, encouraging multidisciplinary collaborations that can harness technology to enhance patient care, improve diagnostic accuracy, and ultimately advance medical science.

\subsection{Future Work}
While this study has made significant strides in leveraging deep learning for PKD detection, several avenues for future research remain to further enhance the accuracy and applicability of the developed models.

One critical area of improvement lies in the utilization of more authentic and diverse human data for training and testing. The success achieved with the mice dataset demonstrates the potential of the MLP algorithm, but its viability can only be fully analyzed with human-specific data. Acquiring a comprehensive dataset that encompasses a wide range of patient profiles, genetic variations, and disease stages is essential, and therefore we aim to collaborate with medical institutions in the future to gather such datasets to further this model into clinical use. Moreover, exploring ways to overcome data scarcity through data augmentation techniques or synthetic data generation methods that accurately simulate human data can provide valuable insights into the robustness of the models.

Furthermore, this study primarily focused on PKD detection, but the potential of deep learning in predicting disease progression warrants exploration. Developing predictive models that can forecast the progression of PKD and identify patients at higher risk of developing complications could significantly impact patient care and treatment strategies.
\section{Acknowledgment}
We would like to thank the University of North Texas for providing us with the resources and support to conduct this research. The invaluable guidance and encouragement from our professors and mentors have been instrumental in shaping the direction and scope of this study. We would also like to acknowledge the National Kidney Foundation for their support and inspiration in conducting this project. Finally, we would also like to thank our families for supporting us throughout our research.
\bibliographystyle{IEEEtran}
\bibliography{Bibliography}

\begin{thebibliography}{99}

\bibitem{Bergmann2018}
Carsten Bergmann, Lisa M. Guay-Woodford, Peter C. Harris, Shigeo Horie, Dorien J. M. Peters, Vicente E. Torres. \textit{Polycystic kidney disease}. Nature Reviews Disease Primers 4(1):50, 2018.

\bibitem{cyst}
Mehrdad Mohammadi Sichani, Reza Safi, Saeid Haghdani, MohammadHatef Khorrami, Farshid Alizadeh, MohammadHossein Izadpanahi. \textit{Does the simple renal cyst treatment improve renal function: A pilot study}. Advanced Biomedical Research 11(1):38, 2022. 

\bibitem{CKD}
Charlotte Gimpel, E. Fred Avni, Luc Breysem, Kathrin Burgmaier, Anna Caroli, Metin Cetiner, Dieter Haffner, Erum A. Hartung, Doris Franke, Jens König, Max C. Liebau, Djalila Mekahli, Albert C. M. Ong, Lars Pape, Andrea Titieni, Roser Torra, Paul J. D. Winyard, Franz Schaefer. \textit{Imaging of Kidney Cysts and Cystic Kidney Diseases in Children: An International Working Group Consensus Statement}. Radiology 290(3): 769-782, 2019.

\bibitem{DL}
Riccardo Miotto, Fei Wang, Shuang Wang, Xiaoqian Jiang, Joel T Dudley. \textit{Deep learning for healthcare: review, opportunities and challenges}. Briefings in Bioinformatics 19(6):1236-1246, 2018.

\bibitem{NN}
Nida Shahid, Tim Rappon, Whitney Berta. \textit{Applications of artificial neural networks in health care organizational decision-making: A scoping review}. PLOS ONE 14(2): e0212356, 2019.

\bibitem{Chen2008}
Wen-Cheng Chen, Yi-Shiuan Tzeng, Hung Li. \textit{Gene expression in early and progression phases of autosomal dominant polycystic kidney disease}. BMC Research Notes 1:131, 2008.

\bibitem{jordon2022synthetic}
James Jordon, Lukasz Szpruch, Florimond Houssiau, Mirko Bottarelli, Giovanni Cherubin, Carsten Maple, Samuel N. Cohen, Adrian Weller. \textit{Synthetic Data -- what, why and how?}. 2022.

\bibitem{Eden2009GOrilla}  
Eran Eden, Roy Navon, Israel Steinfeld, Doron Lipson, Zohar Yakhini. \textit{GOrilla: A Tool For Discovery And Visualization of Enriched GO Terms in Ranked Gene Lists}. BMC Bioinformatics 10:48, 2009.

\bibitem{Lin2013}
Wei-Jye Lin, Stephen R. Salton. \textit{The Regulated Secretory Pathway and Human Disease: Insights from Gene Variants and Single Nucleotide Polymorphisms}. Frontiers in Endocrinology 4:96, 2013.

\bibitem{Jiang2018}
Hong Jiang, Xiaoyu Zhang, Xiao Chen, Pornpun Aramsangtienchai, Zhen Tong, Hening Lin. \textit{Protein Lipidation: Occurrence, Mechanisms, Biological Functions, and Enabling Technologies}. Chemical Reviews 118(3): 919-988, 2018. 

\bibitem{Chittenden2002} 
Thomas Chittenden. \textit{BH3 domains: intracellular death-ligands critical for initiating apoptosis}. Cancer Cell 2(3): 165-166, 2002.

\bibitem{SINGH1998}
ARJUN SINGH, JIAN NI, BHARAT B. AGGARWAL. \textit{Review: Death Domain Receptors and Their Role in Cell Demise}. Journal of Interferon \& Cytokine Research 18(7): 439-450, 1998. 

\bibitem{Panda2023}
Kapil Panda, Anirudh Mazumder. \textit{Predicting Dosage of Immunosuppressant Drugs After Kidney Transplantation Using Machine Learning}. 2023.

\bibitem{aki}
Kapil Panda. \textit{Gene Expression Analysis of Acute Kidney Injury in Kidney Transplants}. Journal of Student Research. 2022.

\bibitem{Perlman2016}
Robert L Perlman. \textit{Mouse models of human disease: An evolutionary perspective}. Evolution, Medicine, and Public Health 2016(1): 170-176, 2016.

\bibitem{Ernst2018}
Peter B. Ernst, Anne-Ruxandra Carvunis. \textit{Of mice, men and immunity: a case for evolutionary systems biology}. Nature Immunology 19(5): 421-425, 2018. 

\bibitem{James2021}
Stefanie James, Chris Harbron, Janice Branson, Mimmi Sundler. \textit{Synthetic data use: exploring use cases to optimise data utility}. Discover Artificial Intelligence 1(1):15, 2021.

\end{thebibliography}

\end{document}